\documentclass[letterpaper, 10 pt, conference]{ieeeconf}  

\IEEEoverridecommandlockouts                              

\usepackage{graphicx}
\usepackage{amsmath, amssymb}
\usepackage{makecell}
\usepackage{multirow}
\usepackage{subfigure}
\usepackage{url}
\usepackage{threeparttable}
\usepackage{cite}
\usepackage{bm}
\usepackage{pdfpages}
\allowdisplaybreaks

\DeclareGraphicsExtensions{.png,.pdf}

\title{\LARGE \bf
Immersive Demonstrations are the Key to Imitation Learning
}

\author{Kelin Li\textsuperscript{$\dagger$}, Digby Chappell\textsuperscript{$\dagger$}, and~Nicolas~Rojas
\thanks{$\dagger$ Authors contributed equally to this work.}
\thanks{Kelin Li was partly supported by China Scholarship Council and Dyson School of Design Engineering, Imperial College London. Digby Chappell was supported by the UKRI CDT in AI for Healthcare \protect \url{http://ai4health.io} (Grant No. EP/S023283/1).}
\thanks{Kelin Li, Digby Chappell, and Nicolas Rojas are with the REDS Lab, Dyson School of Design Engineering, Imperial College London, 25 Exhibition Road, London, SW7 2DB, UK
{\tt\small (k.li20, d.chappell19, n.rojas)@imperial.ac.uk}}
}

\begin{document}

\maketitle
\thispagestyle{empty}
\pagestyle{empty}

\begin{abstract}
Achieving successful robotic manipulation is an essential step towards robots being widely used in industry and home settings. Recently, many learning-based methods have been proposed to tackle this challenge, with imitation learning showing great promise. However, imperfect demonstrations and a lack of feedback from teleoperation systems may lead to poor or even unsafe results. In this work we explore the effect of demonstrator force feedback on imitation learning, using a feedback glove and a robot arm to render fingertip-level and palm-level forces, respectively. 10 participants recorded 5 demonstrations of a pick-and-place task with 3 grippers, under conditions with no force feedback, fingertip force feedback, and fingertip and palm force feedback. Results show that force feedback significantly reduces demonstrator fingertip and palm forces, leads to a lower variation in demonstrator forces, and recorded trajectories that a quicker to execute. Using behavioral cloning, we find that agents trained to imitate these trajectories mirror these benefits, even though agents have no force data shown to them during training. We conclude that immersive demonstrations, achieved with force feedback, may be the key to unlocking safer, quicker to execute dexterous manipulation policies.
\end{abstract}

\section{Introduction}
\label{sec:introduction}


Robotic manipulation is one of the most important capabilities that robotic hands must have in order to be used for household tasks. The traditional method to endow robots with manipulation skills is by hard coding motions and grasps, which requires different programs for different tasks and environments. In recent years, Artificial Intelligence (AI) is becoming more popular in robotic manipulation \cite{Zhang2018DeepTeleoperation,Mandlekar2020Human-in-the-LoopTeleoperation,Edmonds2017FeelingBottles,Chen2022ARe-Orientation,Kuklinski2014TeleoperationControl,Johns2021Coarse-to-FineDemonstration}. Using these techniques, the robot can learn a manipulation policy from experiences collected either by itself or from demonstrations. The learned policy enables the robot to choose an action at each timestep after perceiving the current state of its environment \cite{Argall2009ADemonstration}, and if the robot has been exposed to a sufficiently large sample of states and actions, then the learnt policy will be robust and accurate. Imitation learning, also known as Learning from Demonstration (LfD) is one of the most popular AI methods to achieve this.

\begin{figure}[t!]
    \centering
    \includegraphics[width=\columnwidth]{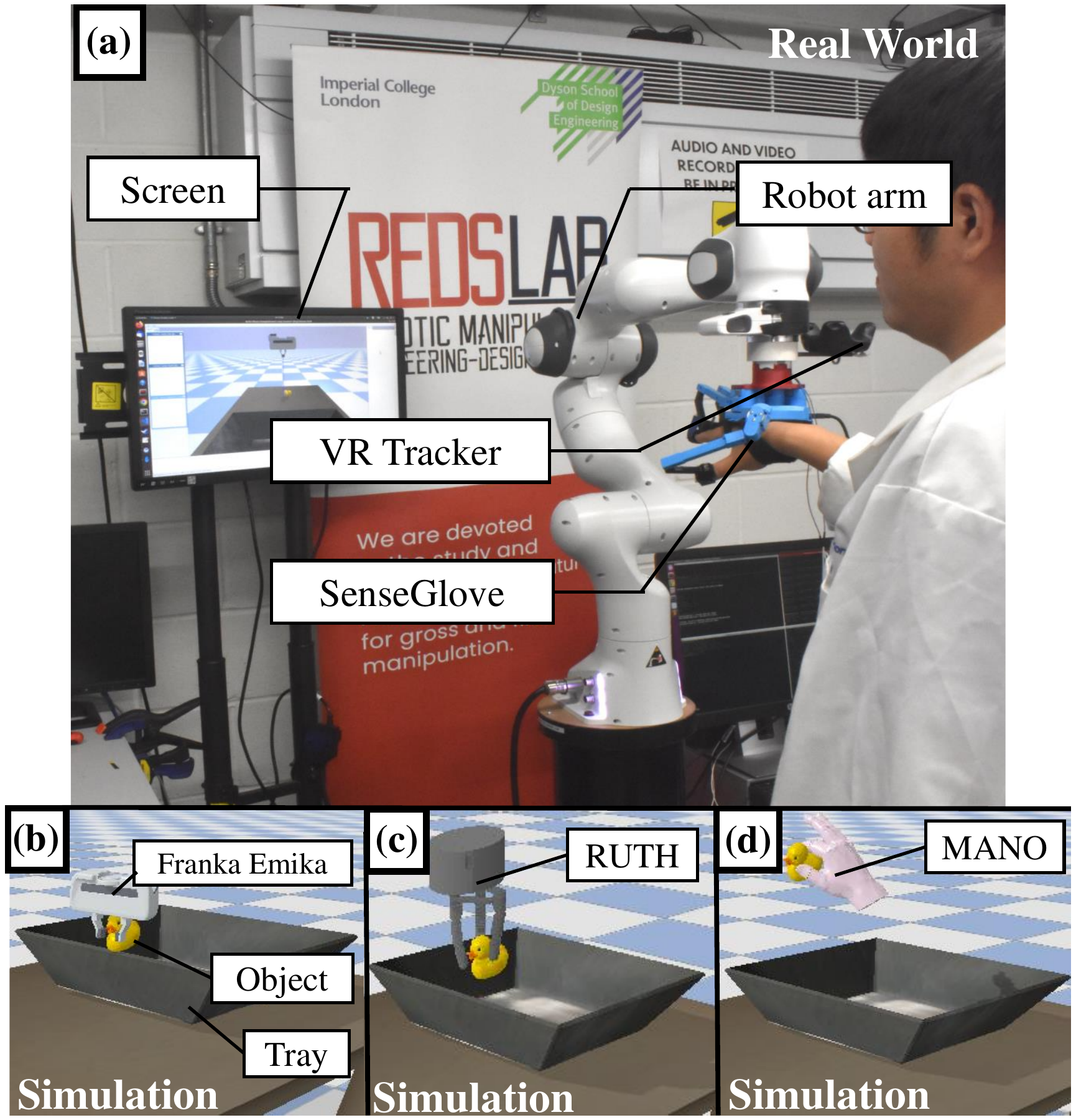}
    \caption{Summary diagram of the immersive demonstration platform used in this work, which utilizes a robot arm and a haptic feedback glove, shown in (a) to provide sensory feedback from simulated Franka Emika (b), RUTH (c), and MANO (d) gripper/hand.}
    \label{fig:main}
    \vspace{-4mm}
\end{figure}

Naturally, quality of the demonstration shown to the robot greatly affects the results of imitation learning \cite{Wu2019ImitationDemonstration}. Imperfect demonstrations will result in an imperfect imitation policy, and therefore it is essential to have a high quality demonstration collection process. Broadly speaking, there are two categories of demonstration collection techniques: indirect demonstration and direct demonstration \cite{Fang2019SurveyManipulation}. For indirect demonstration, the expert does not interact with the robot; instead, the expert's own trajectory of state and actions are recorded with a visual or wearable system, and are retargeted to the robot. For direct demonstration, robot is guided by the expert to carry out the task, allowing the robot to learn from its own motion data directly. While direct demonstration methods minimize issues with retargeting motion data, they inherently prevent the demonstrator from performing a truly expert demonstration. Further to this, many indirect demonstration frameworks use virtual reality platforms to enable precise monitoring of the state of the environment and the demonstrator~\cite{Li2022IGibsonTasks, Shen2021IGibsonScenes}. However, the absence of many sensory feedback mechanisms (e.g. contact force, fingertip tactile) in such systems means that demonstrations are imperfect \cite{Jing2020ReinforcementGuidance}, and this may negatively impact imitation learning results. Thus, improving demonstration platforms with sensory feedback mechanisms may be a fruitful avenue of research for improving imitation learning.

In this paper, we explore the effect of demonstrator sensory feedback on imitation learning. By utilizing a haptic feedback glove and a robot arm, we are able to render fingertip and palm-level force feedback to the demonstrator, allowing the demonstrator to feel interactions during manipulation tasks. We use this system, summarized in Fig.~\ref{fig:main}, to collect expert demonstrations for a pick and place manipulation task using three simulated grippers: the 1 degree-of-freedom (DoF) Franka Emika Hand \cite{Haddadin2022TheEducation}, the 3 DoF RUTH gripper \cite{Lu2021SystematicGripper}, and the 20 DoF MANO hand \cite{Romero2017EmbodiedHands}. Experimental results indicate that demonstration collected with fingertip feedback significantly reduces the gripper force that the trained policy uses, while maintaining a high task success rate. Furthermore, with palm-level force feedback, the force applied to each gripper to control their pose is greatly reduced.

This paper is organized as follows. In Section \ref{sec:related_work}, related works about robotic manipulation and the methods proposed to address it are reviewed. We show how we formulate the fingertip force and palm force from the simulation environment in Section \ref{sec:methods}. The imitation learning we adopt in this paper is also explained in this section. We introduce the experimental setup we use for this paper in Section \ref{sec:experimental_setup}. Experimental results and some discussions are done in Section \ref{sec:results_discussion}. Finally, we conclude this paper and raise the future plans to improve the current work in Section \ref{sec:conclusions}.

\section{Related Work}

\label{sec:related_work}
\subsection{Robotic Manipulation}\label{sec:manipulation}
Robotic manipulation, including grasping, relocation, and reorientation, is a crucial capability for robotics \cite{Billard2019TrendsManipulation}. A large amount of research on the subject has been performed, with works roughly falling into the categories of robot arm manipulations and dexterous manipulations. Both traditional control methods and learning methods have been widely employed for robot arm manipulations. For example, traditional control methods were applied to robot arm systems performing machining task \cite{Li2017ForceFiltering} and coffee machine tasks \cite{Mandlekar2020Human-in-the-LoopTeleoperation}. Dexterous manipulation skills, on the other hand, are typically achieved with learning-based methods, such as deep learning to grasp complex objects \cite{Li2022EfficientGrasp:Hands}, or learning from human demonstration to perform unscrewing tasks \cite{Edmonds2017FeelingBottles}, door opening \cite{Zhu2019DexterousLow-Cost}, and in-hand manipulation \cite{Chen2022ARe-Orientation}.

\subsection{AI Methods for Robotic Manipulation}
\label{sec:ai_methods}
In recent years, researchers have focused on applying machine learning methods to robotic manipulation problems. One of the most popular methods is reinforcement learning (RL). With a well formulated reward function, RL can automatically learn a policy which maps states to actions \cite{RichardS.Sutton2018ReinforcementIntroduction}, allowing it to be widely applied to robotic manipulation tasks. Many works have focused on learning policies directly from large dimension inputs, often taking RGB images as observations and robot behaviors as actions \cite{Gu2017DeepUpdates,Zhan2021AManipulation}. An open-source framework named SURREAL \cite{Fan2018SURREAL:Benchmark} was proposed to accelerate the deep reinforcement learning, which shows high scalability and strong results in the applications. RL has been successfully used to train a dexterous robotic hand to carry out tasks, such as valve rotation and box flipping \cite{Zhu2019DexterousLow-Cost}, but for many tasks, it is too difficult to define a suitable reward function. Furthermore, for high dimension state and action space problems such as robotic manipulation, it is inefficient to learn from scratch, rather than utilizing prior knowledge to accelerate learning. Therefore, improving methods that can learn a policy from an expert's demonstration is an appealing topic of research for robotic manipulation.

Imitation learning is a promising paradigm that can learn a policy from demonstrations which are provided by an expert. Incorporating imitation learning into robotic manipulation has gained popularity in recent years, and has proven to be particularly useful. For example, in \cite{Chen2022ARe-Orientation}, a model-free framework was proposed for in-hand object reorientation, which learnt a policy from a pre-trained robotic expert and can deal with unforeseen objects. An alternative way to imitate experts' behaviors is learn from visual sensing \cite{Johns2021Coarse-to-FineDemonstration, Sermanet2018Time-ContrastiveVideo}, where raw video demonstrations can be taken for imitation and used to create a reward function for RL. In \cite{Johns2021Coarse-to-FineDemonstration}, a method that can learn a novel robot manipulation task from a single human demonstration was proposed, which makes visual imitation learning more efficient. However, it is sometimes difficult to map from visual data to robot motion, for example in the presence of occlusion and noise, and visual data may also not be able to infer certain inputs such as force or tactile data. A more direct way of collecting demonstrations is to learn from an expert robot motion directly. Teleoperation \cite{Mandlekar2020Human-in-the-LoopTeleoperation, Kuklinski2014TeleoperationControl} and virtual reality (VR) \cite{Zhang2018DeepTeleoperation} are two promising ways to achieve this.

\subsection{Demonstration Acquisition for Imitation Learning}
\label{sec:demonstration}
Demonstration acquisition is a crucial step in imitation learning; imperfect demonstrations will lead to an unsuccessful learning result. Methods for collecting demonstrations for imitation learning broadly fall into two categories: indirect demonstrations and direct demonstrations \cite{Fang2019SurveyManipulation}.

When performing indirect demonstrations, the expert does not need to contact the robot; the demonstration is inferred, mapped, or retargeted to the robot after it is collected. Most indirect demonstrations are carried out by visual systems, showing great results using unlabelled videos for pouring tasks~\cite{Sermanet2018Time-ContrastiveVideo}, assembly tasks~\cite{Wan2017TeachingVision}, and target reaching~\cite{Johns2021Coarse-to-FineDemonstration}. Another method of performing indirect demonstration is to record demonstrator motion using alternative systems such as wearable devices, such as data gloves. In \cite{Edmonds2017FeelingBottles}, a human expert demonstrated an unscrewing bottle cap task while wearing a dataglove, recording contact force as well as motion data. Although indirect demonstrations are easy to perform, a post-demonstration mapping or learning is required for demonstrations to be used by a robot, and this mapping cannot be guaranteed to align well with the joint capabilities of the robot.

\begin{figure*}
    \vspace{4mm}
    \centering
    \includegraphics[width=0.85\textwidth]{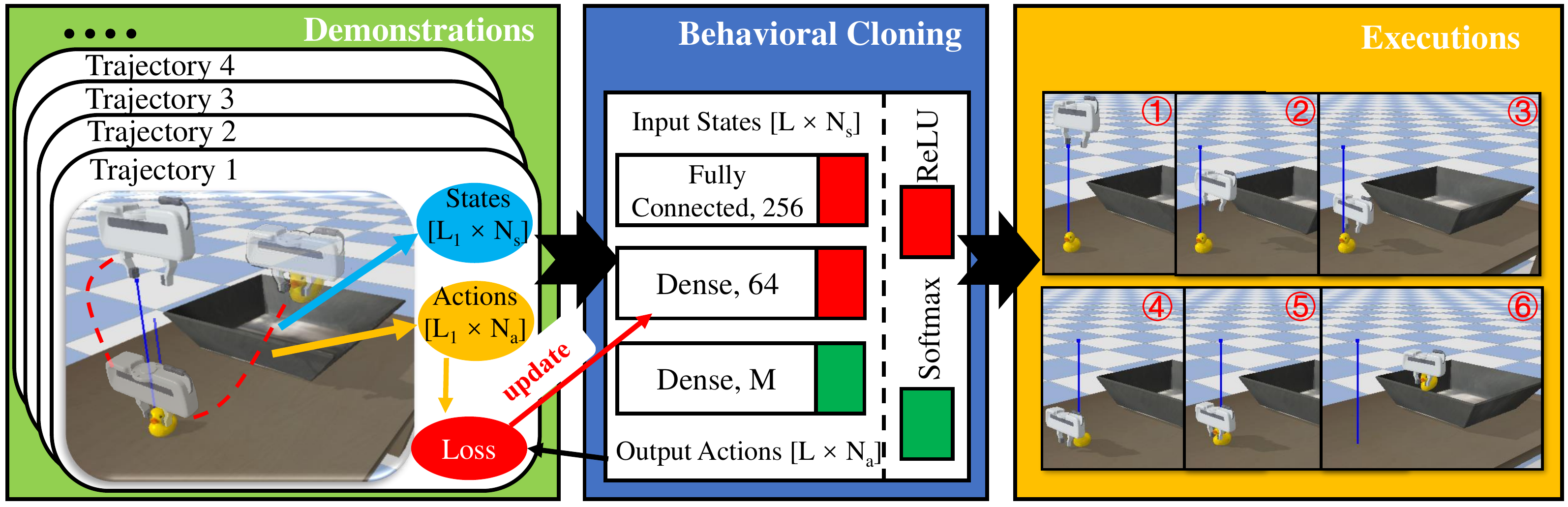}
    \caption{Overview of the behavioral cloning process, here shown with the Franka Emika gripper.}
    \label{fig:BC}
    \vspace{-4mm}
\end{figure*}

Direct demonstrations, on the other hand, directly control the robot to perform the specific task. In many works, this involves guiding the robot externally via kinesthetic teaching~\cite{Caccavale2019KinestheticInteraction, Kormushev2011ImitationInput}, however it is not clear exactly how well these demonstrations truly reflect those of an expert. Alternatively, teleoperation systems can be used, allowing the direct, non-contact control of the robot in an intuitive way. Generally, systems using remote teleoperation have shown promising results on a range of manipulation tasks~\cite{Mandlekar2020Human-in-the-LoopTeleoperation, Kuklinski2014TeleoperationControl}. A more immersive, intuitive method of teleoperation is virtual telepresence, allowing the demonstrator to obtain a first person view of the robot being operated using a virtual reality headset. This has been successfully applied in the case of \cite{Zhang2018DeepTeleoperation}, where virtual telepresence demonstrations were used with deep imitation learning to allow a robot to complete reorientation and grasping tasks. A similar demonstration method also using virtual reality is to use a simulated robot within the virtual environment. This is appealing because it allows every detail of the simulated environment to be collected as part of the demonstration, such as contact forces and joint forces, which are otherwise difficult to monitor~\cite{Kormushev2011ImitationInput}. However, in many teleoperation platforms, particularly those which use virtual telepresence~\cite{Zhang2018DeepTeleoperation}, virtual reality~\cite{Shen2021IGibsonScenes, Li2022IGibsonTasks, Srivastava2022BEHAVIOR:Environments}, and simulation, haptic feedback is not rendered to the demonstrator, meaning the teleoperated demonstrations are imperfect and, in some cases, potentially unsafe.

\section{Methods}
\label{sec:methods}
\subsection{Fingertip Force Feedback}

As mentioned in Section \ref{sec:demonstration}, haptic feedback is important for a human expert to perform high-quality demonstrations. Although rich contact information is available from simulation, rendering this to the demonstrator is limited by existing feedback technology. In this work, we use a force feedback glove, which applies resistive forces in the flexion direction at the tip of each finger. In order to approximate numerous contacts as a single force on the fingertip, we consider the average joint torques applied to the joints of the demonstrator's finger. The flexion joint torques $\tau_{i,\,j}$ at joint $j$ of finger $i$ caused by applying a fingertip force $f_i$ at a distance $d$ from the distal joint of the finger are equal to:
\begin{align}
    \label{eq:torque2force1}
    \tau_{i,\,1}& = d f_i,\\
    \label{eq:torque2force2}
    \tau_{i,\,2}& = l_{i,\,2}\cos(q_{i,\,3})f_i + \tau_{i,\,1},\\
    \label{eq:torque2force3}
    \tau_{i,\,3}& = l_1\cos(q_{i,\,2} + q_{i,\,3})f_i + \tau_{i,\,2},
\end{align}
where $l_{i,\,j}$ is the length of the subsequent phalanx of joint $j$, and $q_{i,\,j}$ is the joint angle of joint $j$. The fingertip force is then computed to minimize the squared error between the average torque, $\hat{\tau}_{i}$, of the demonstrator's finger and the simulated finger or gripper $\hat{\tau}_{i}^{(s)}$
\begin{equation}
\label{eq:torque2forc4}
    \min_{f_i} ||\hat{\tau}_{i}^{(s)} - \hat{\tau}_{i}||^2_2.
\end{equation}

\subsection{Palm Force Feedback}

Although fingertip force feedback allows the demonstrator to feel in-hand forces, the vast majority of feedback gloves only apply forces relative to the palm of the user; the glove itself is fixed to the rear of the palm. In this work we consider how to extend force feedback applied to the demonstrator to also render the external forces acting on the hand. This is particularly important when interacting with static objects (e.g. a desk) in the environment, or when feeling the weight of objects. Using a similar method to \cite{Chappell2022VirtualInteraction}, the base link of the simulated hand or gripper can be controlled with a $6$-DoF proportional-derivative (PD) controller:
\begin{align}
    \bm{F} &= K_{p,x}(\bm{x} - \bm{x}_{ref}) - K_{d,x}\dot{\bm{x}},
    \label{eq:pos_controller}\\
    \bm{\mathcal{T}} &= K_{d,r}(\bm{\omega} - \bm{\omega}_{ref} ),
    \label{eq:rot_controller}
\end{align}
where $\bm{x}$ and $\bm{\omega}$ are the position and angular velocity of the base link, respectively, and $\bm{x}_{ref}$ and $\bm{\omega}_{ref}$ are the position and angular velocity of the real world hand tracker, respectively. The resultant wrench applied by the PD controller at the simulated base link can then be rendered as the wrench applied by the end effector of a robot arm to the demonstrator's hand in order to allow the demonstrator to `feel' the external, out-of-hand forces applied to them.

\begin{figure}[t!]
    \centering
    \vspace{4mm}
    \includegraphics[width=0.95\columnwidth]{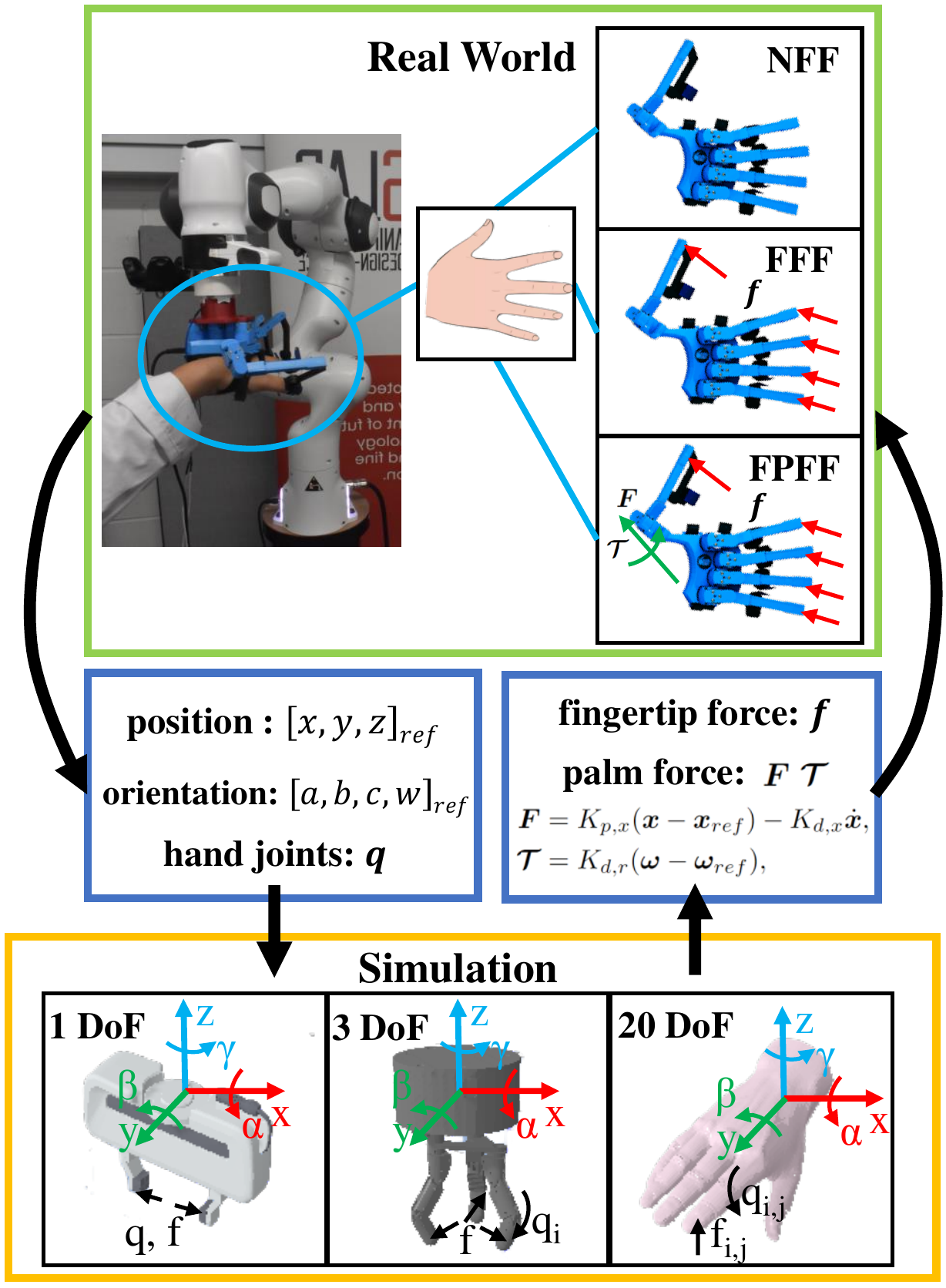}
    \caption{Force feedback in the immersive virtual teleoperation system. Palm pose (position and orientation) and finger joint positions are captured using a SenseGlove and mapped to desired positions for the simulated grippers. Joint forces from the simulation are mapped back to fingertip forces to apply with the SenseGlove and an end-effector wrench to apply with a Franka Emika robot arm to the palm.}
    \label{fig:setup}
    \vspace{-4mm}
\end{figure}

\subsection{Imitation Learning}
\label{subsection:learning}

With collected demonstrations, we train a policy using imitation learning. In this work, a simple Behavioral Cloning (BC) baseline is adopted to imitate the demonstrator's behavior, summarized in Fig.~\ref{fig:BC}. We use three fully connected layers to map the observed robot states to demonstrated actions. All networks were trained by minimizing the mean square error between the demonstration action and the agent's action:
\begin{equation}
\begin{split}
L=\frac{1}{N}\sum\nolimits_{i=1}^{N_a}({^s\hat{a}_i}-{^sa_i})^2,
\end{split}
\end{equation}
where $N$ represents the dimension of the action space. $^s\hat{a}_i$ is the $i_{th}$ agent's action respective to a specific observed state $s$. And ${^sa}_i$ is the ground truth demonstration action. In this work, we use a simple representation where the agent can only perceive its own controllable internal state, corresponding to the position of each degree of freedom. The dimensions of the action space and state space are 7, 9 and 26 for Franka Emika Hand, RUTH hand and MANO hand, respectively (6 DoF palm pose + gripper DoF). Before the demonstrations are fed into the networks, all the data are normalized using min-max normalization:
\begin{equation}
\begin{split}
x'=\frac{x-x_{min}}{x_{max}-x_{min}},
\end{split}
\end{equation}
where $x'$ is normalized and $x$ is original demonstration data.

\begin{figure}[t!]
\centering
\vspace{4mm}
\subfigure[Fingertip force]{\label{fig:fingertip_demo}\includegraphics[width=0.9\columnwidth]{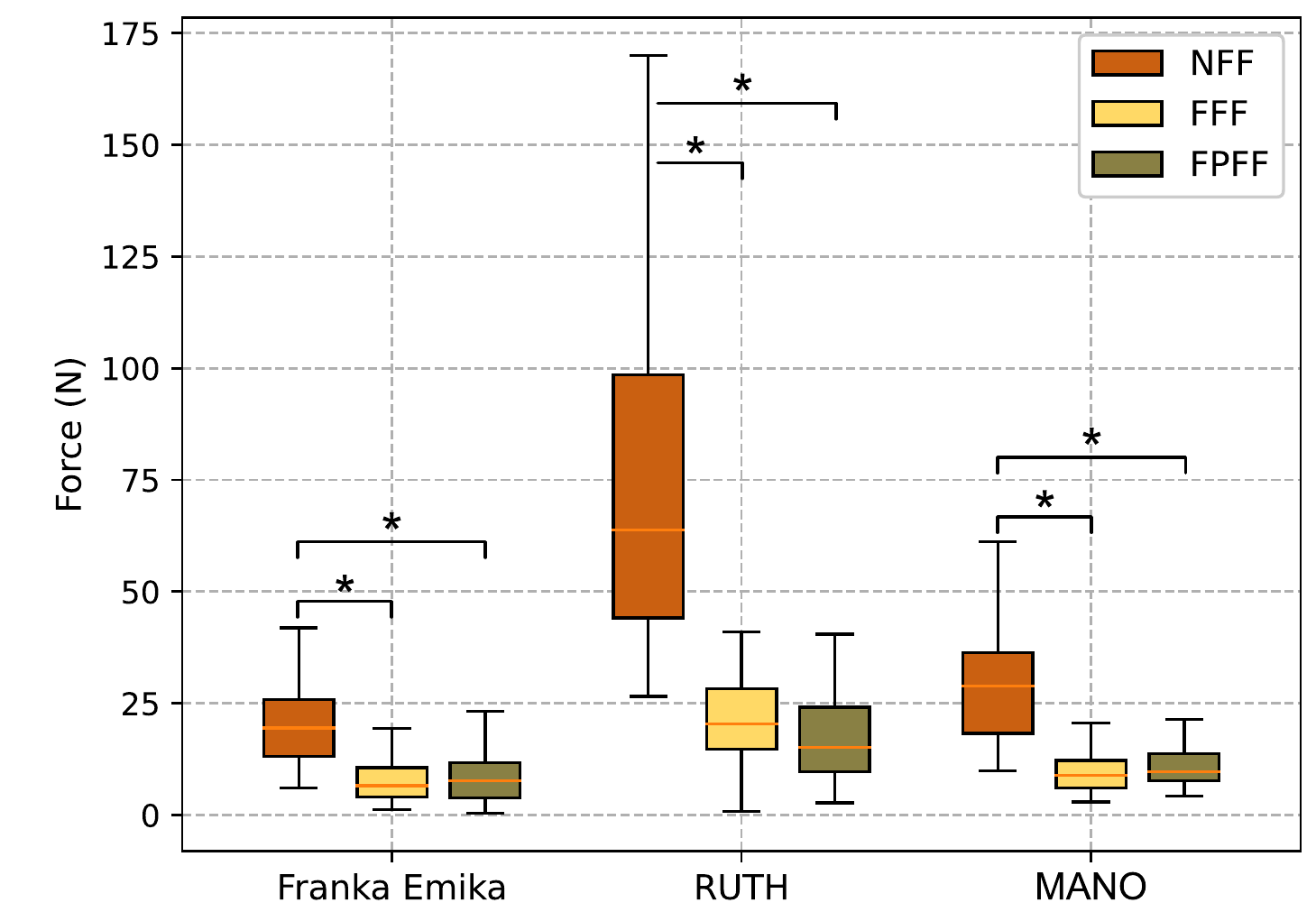}}
\subfigure[Palm force]{\label{fig:palm_demo}\includegraphics[width=0.9\columnwidth]{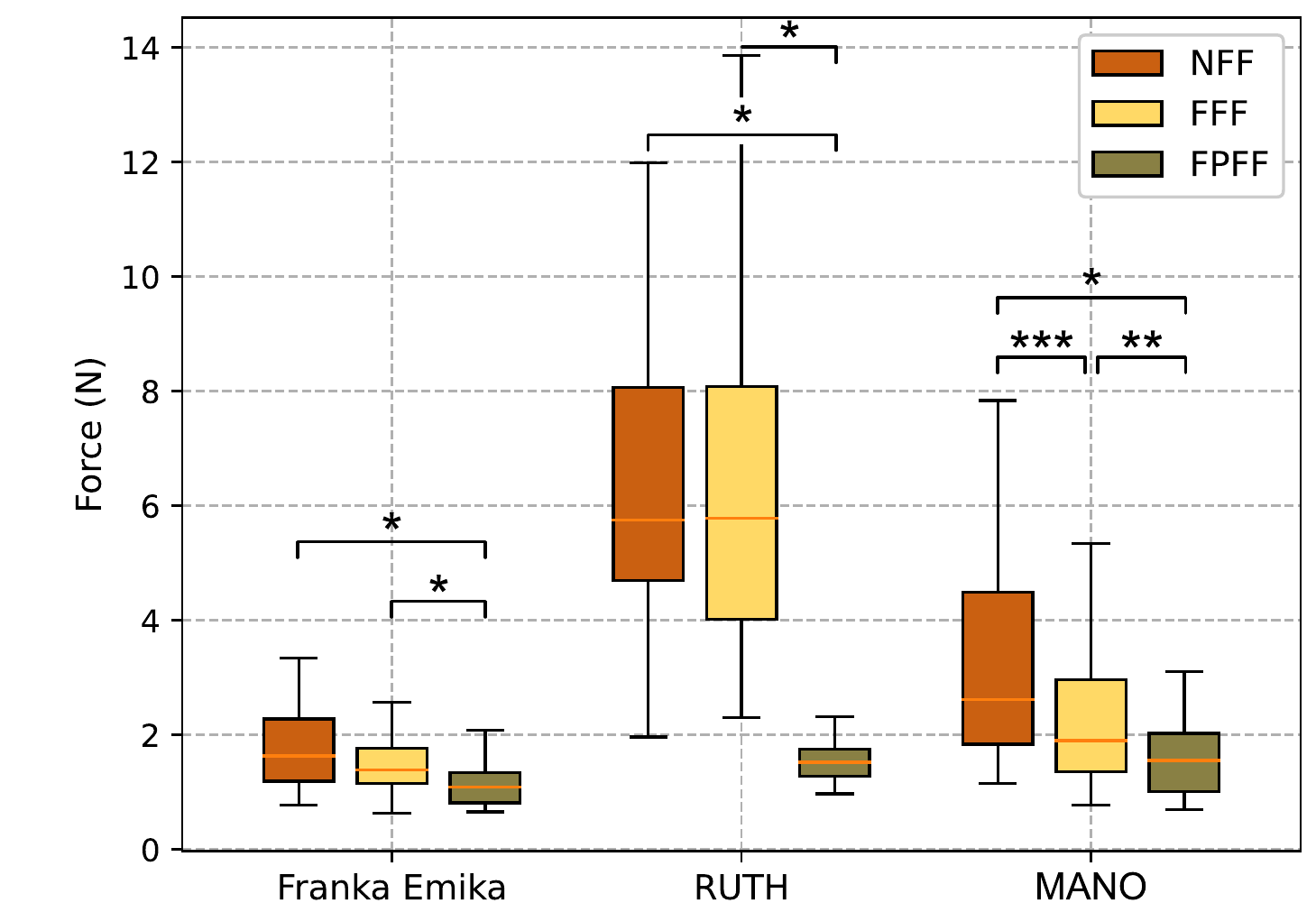}}\\
\vspace{-2mm}
\caption{Box plots of demonstration (a) fingertip forces and (b) palm forces for no force feedback (NFF), fingertip force feedback (FFF), and fingertip and palm force feedback (FPFF). Results of student's $t$-test, *$p < 0.001$, **$p < 0.01$, ***$p<0.05$.}
\label{fig:force_demo}
\vspace{-4mm}
\end{figure}

\section{Experimental Setup}
\label{sec:experimental_setup}
To implement the force feedback methods described previously, a SenseGlove force feedback glove is used to provide resistive forces to the demonstrator's fingertips, and a Franka Emika robot arm to provide palm force feedback, both shown in Fig.~\ref{fig:setup}. Resistive forces are translated to a pulse-width modulation (PWM) signal to send to the force feedback glove. The duty cycle percentage of the signal is computed by empirically fitting a quadratic to measured force outputs from each resistive tendon:
\begin{equation}
\label{eq:pwm}
    \text{\% duty cycle }i = \sqrt{(f_i - b) / a}
\end{equation}
where $a = 1.72\times10^{-3}$, and $b = 2.57$. The palm force feedback wrench, calculated using Eq. (\ref{eq:pos_controller}) and Eq. (\ref{eq:rot_controller}), is applied by the end effector of the robot arm to the back of the demonstrator's hand by robot arm.

\begin{figure*}[t]
\centering    
\vspace{4mm}
\subfigure[Franka Emika]{\label{fig:franka_finger}\includegraphics[width=0.33\textwidth]{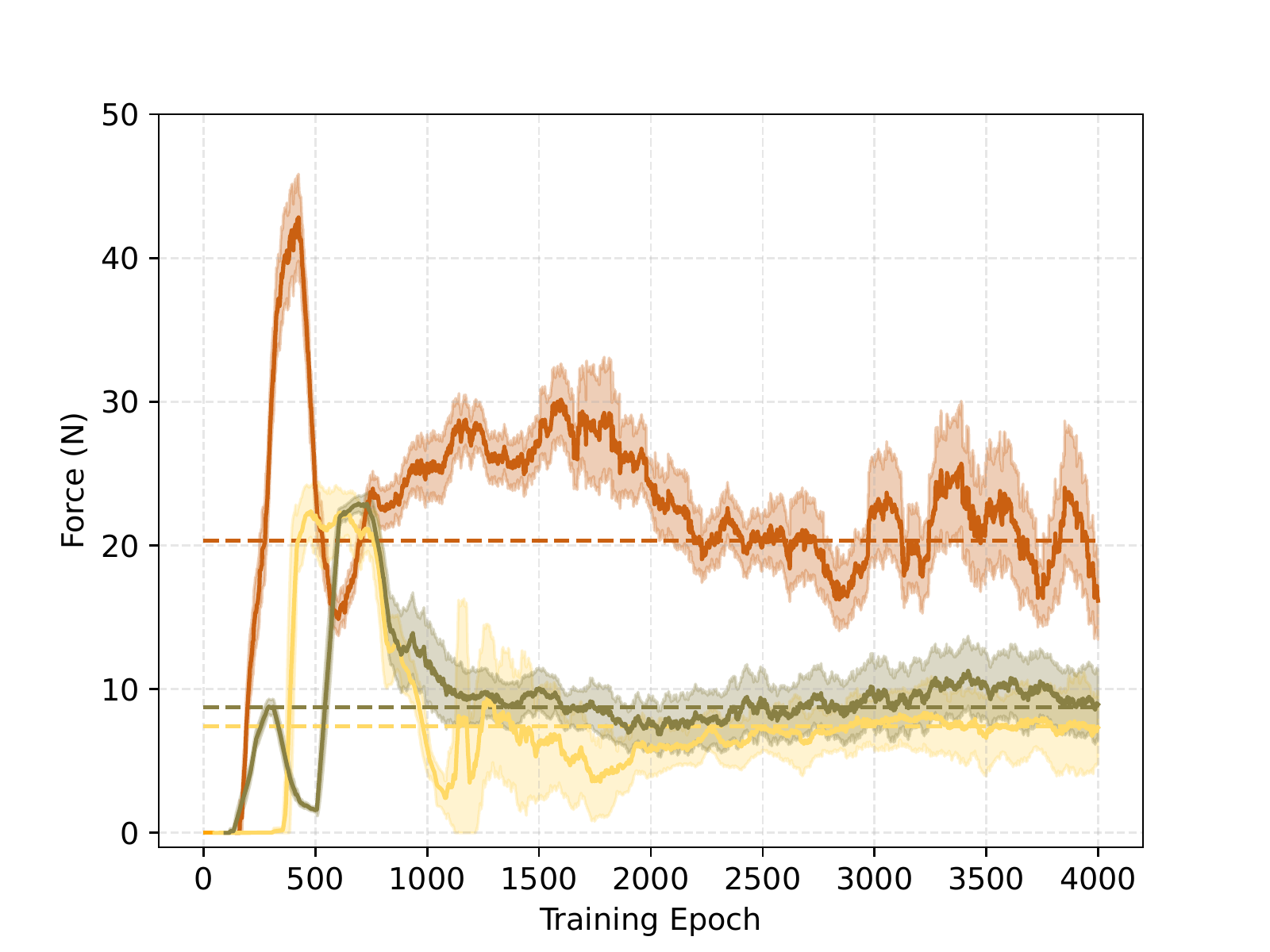}}
\hspace{-0.01\textwidth}
\subfigure[RUTH]{\label{fig:ruth_finger}\includegraphics[width=0.33\textwidth]{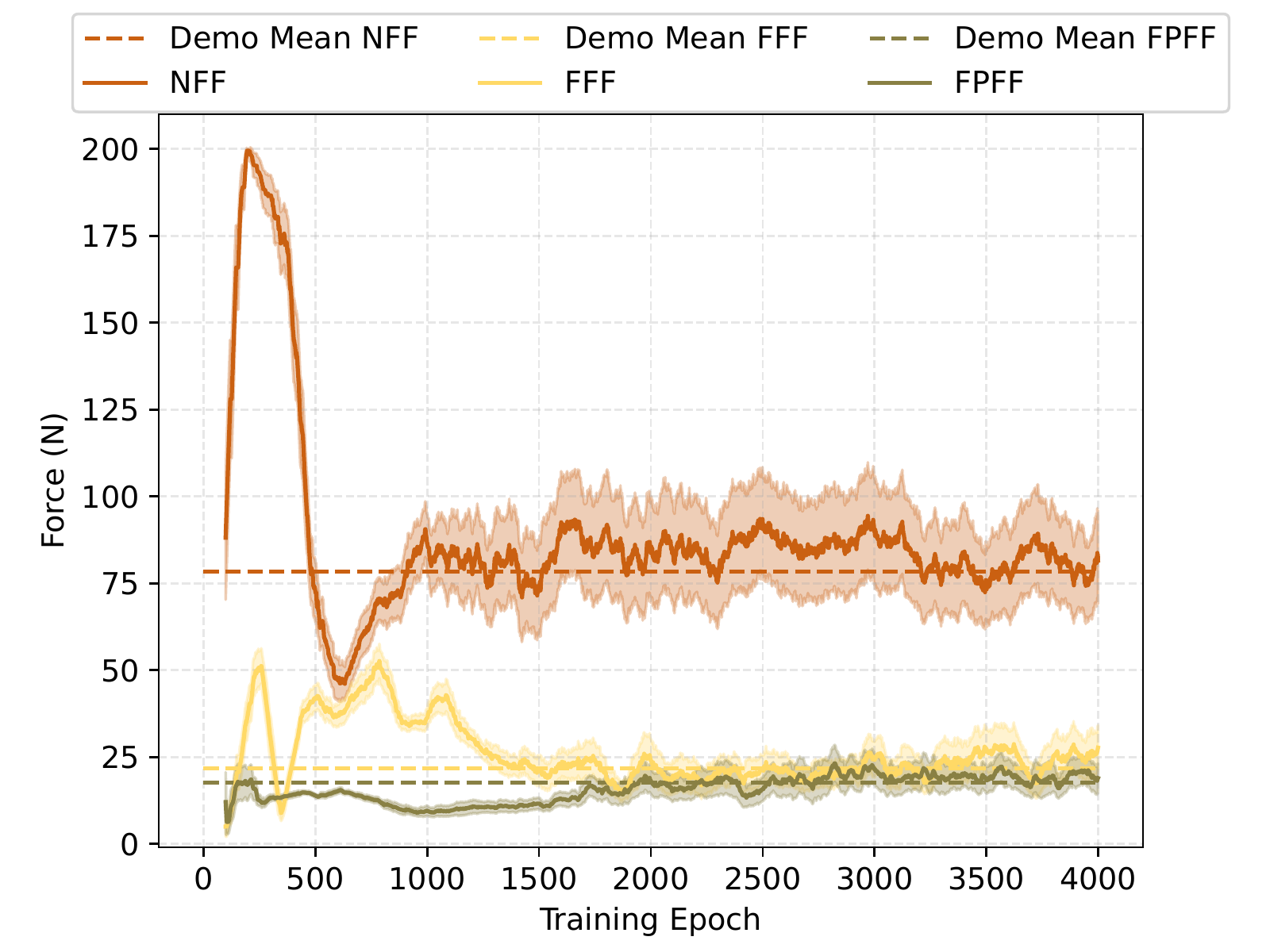}}
\hspace{-0.01\textwidth}
\subfigure[MANO]{\label{fig:mano_finger}\includegraphics[width=0.33\textwidth]{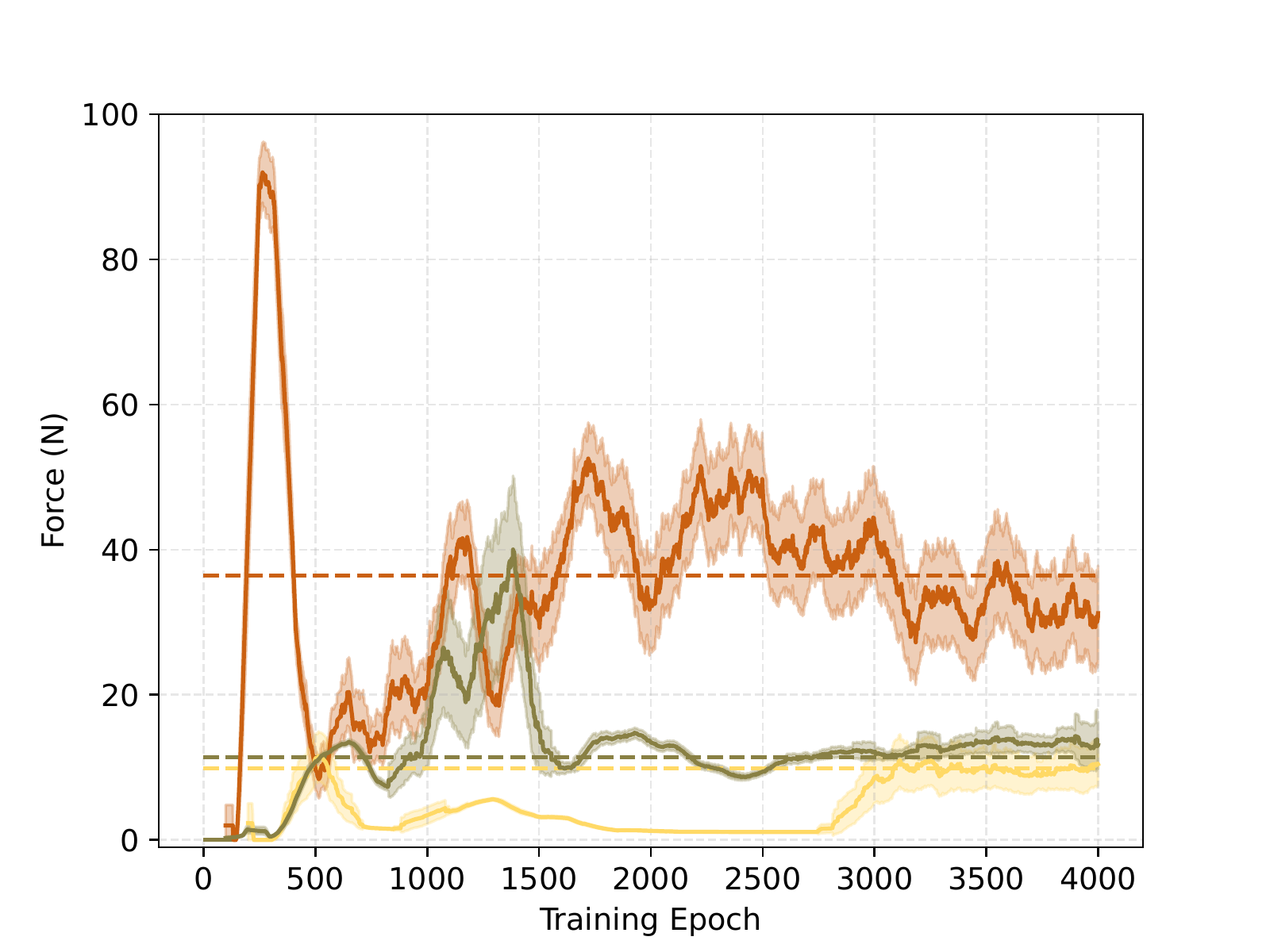}}
\vspace{-4mm}
\caption{Mean fingertip force during the training of imitation learning agents. Rolling window of $100$ epochs is used, $\pm1$ standard deviation is shaded. Dashed lines indicate mean fingertip force from demonstrations.}
\label{fig:fingertip_train}
\vspace{-4mm}
\end{figure*}

The reference position $[x,y,z]_{ref}$ and reference orientation $[a,b,c,w]_{ref}$ of the palm are measured by a VR tracker, which is fixed relative to the palm of the demonstrator. Reference joint positions of the human hand \bm{$q$} are measured using the SenseGlove's proprietary inverse kinematics solver. For the MANO hand, the human hand joint positions are directly mapped to the reference joint positions of the MANO hand and feedback forces directly mapped to the human hand. For the RUTH hand, the flexion joint angles of the most proximal joints of the thumb, index, and middle fingers of the human hand are mapped to the three controllable degrees of freedom of the RUTH hand, with fingertip force feedback from the three fingers of the RUTH hand also being applied to these fingers. For the Franka Emika gripper, the most proximal joint of the index finger is used as a reference for the single controllable degree of freedom, and force feedback rendered to the index finger. Participants were able to view the simulated environment from a fixed perspective on a screen placed at head height approximately 1.5~m in front of them. This increases task difficulty; without being able to alter the viewpoint of the camera, it is possible for the simulated gripper to occlude the object being grasped.

To quantify the effect of force feedback on imitation learning, we consider three conditions: no force feedback (NFF), fingertip force feedback (FFF), and fingertip and palm force feedback (FPFF). In this work, we collect demonstration for a simple pick and place task, defined as grasping a rubber duck and placing it into a tray. Participants are asked to repeatedly perform this task until 5 successful attempts are recorded for each condition, for each gripper. 10 participants were recruited to this study, giving 50 demonstration trajectories for each of the $9$ learning tasks (450 total demonstrations). This study was issued a favorable opinion by the Imperial College London Science Engineering Technology Research Ethics Committee (SETREC), study number 20IC6125, informed consent was obtained from each participant.

\begin{figure*}[t]
\centering
\vspace{4mm}
\subfigure[Franka Emika]{\label{fig:franka_palm}\includegraphics[width=0.33\textwidth]{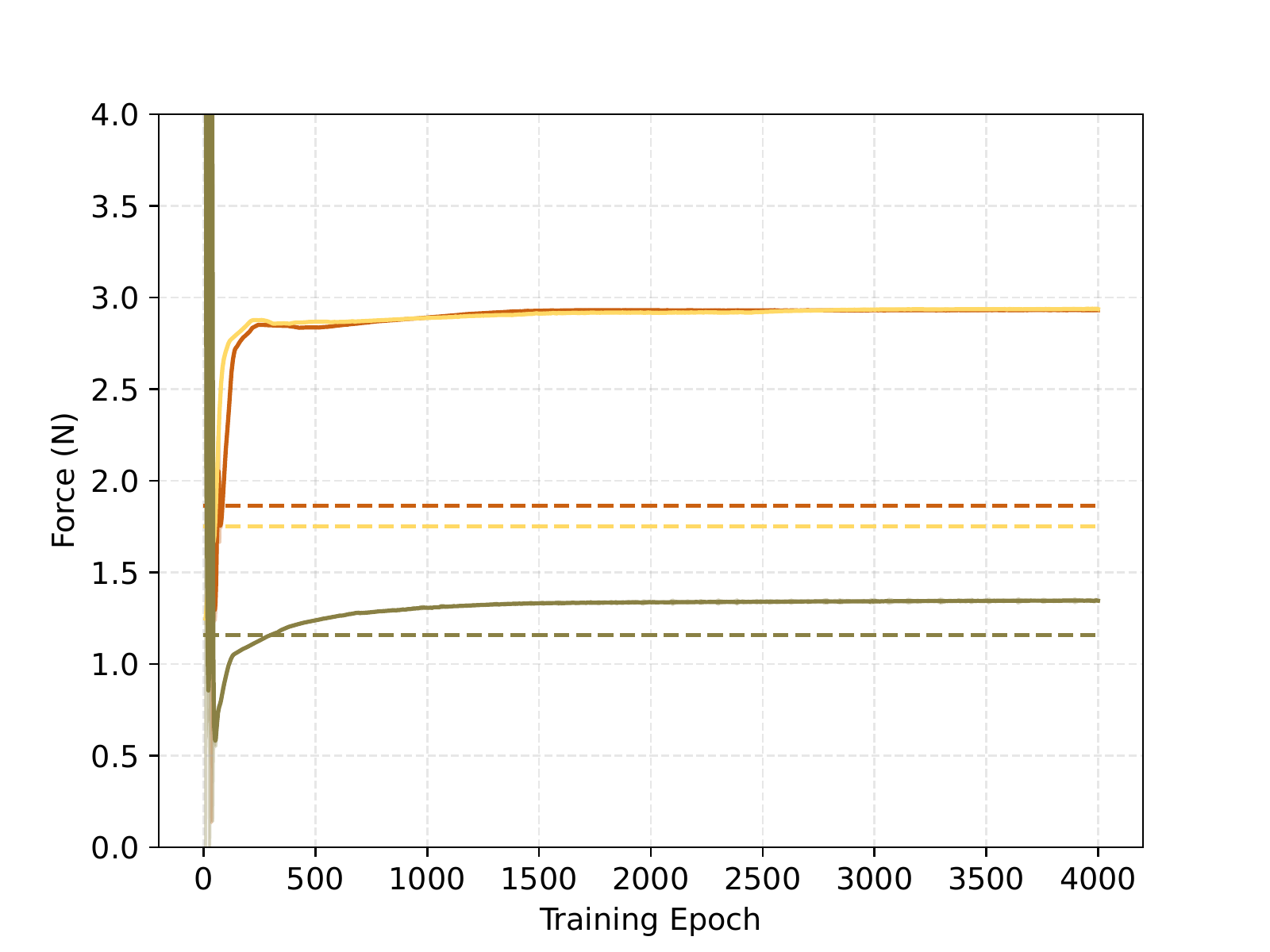}}
\hspace{-0.01\textwidth}
\subfigure[RUTH]{\label{fig:ruth_palm}\includegraphics[width=0.33\textwidth]{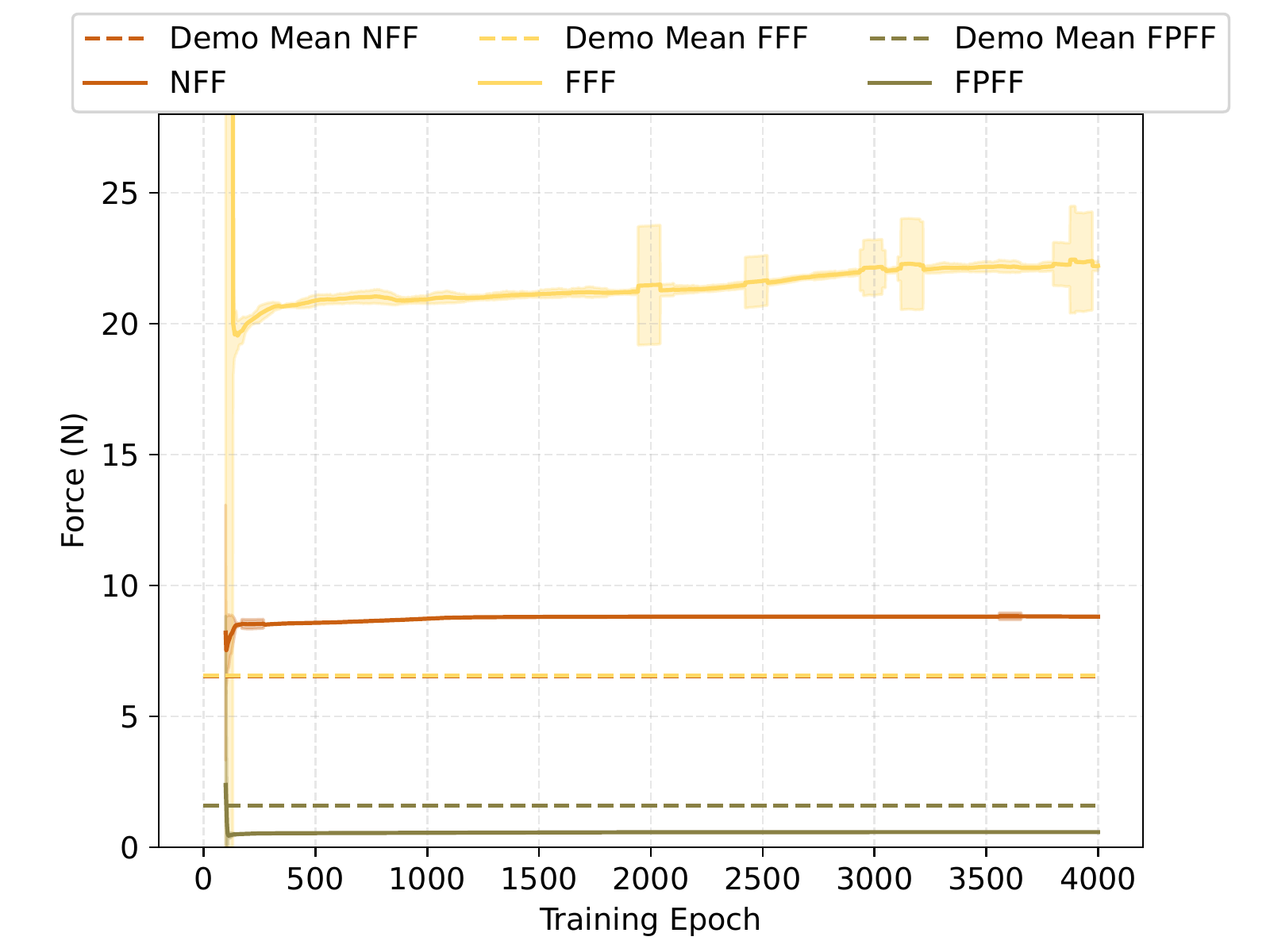}}
\hspace{-0.01\textwidth}
\subfigure[MANO]{\label{fig:mano_palm}\includegraphics[width=0.33\textwidth]{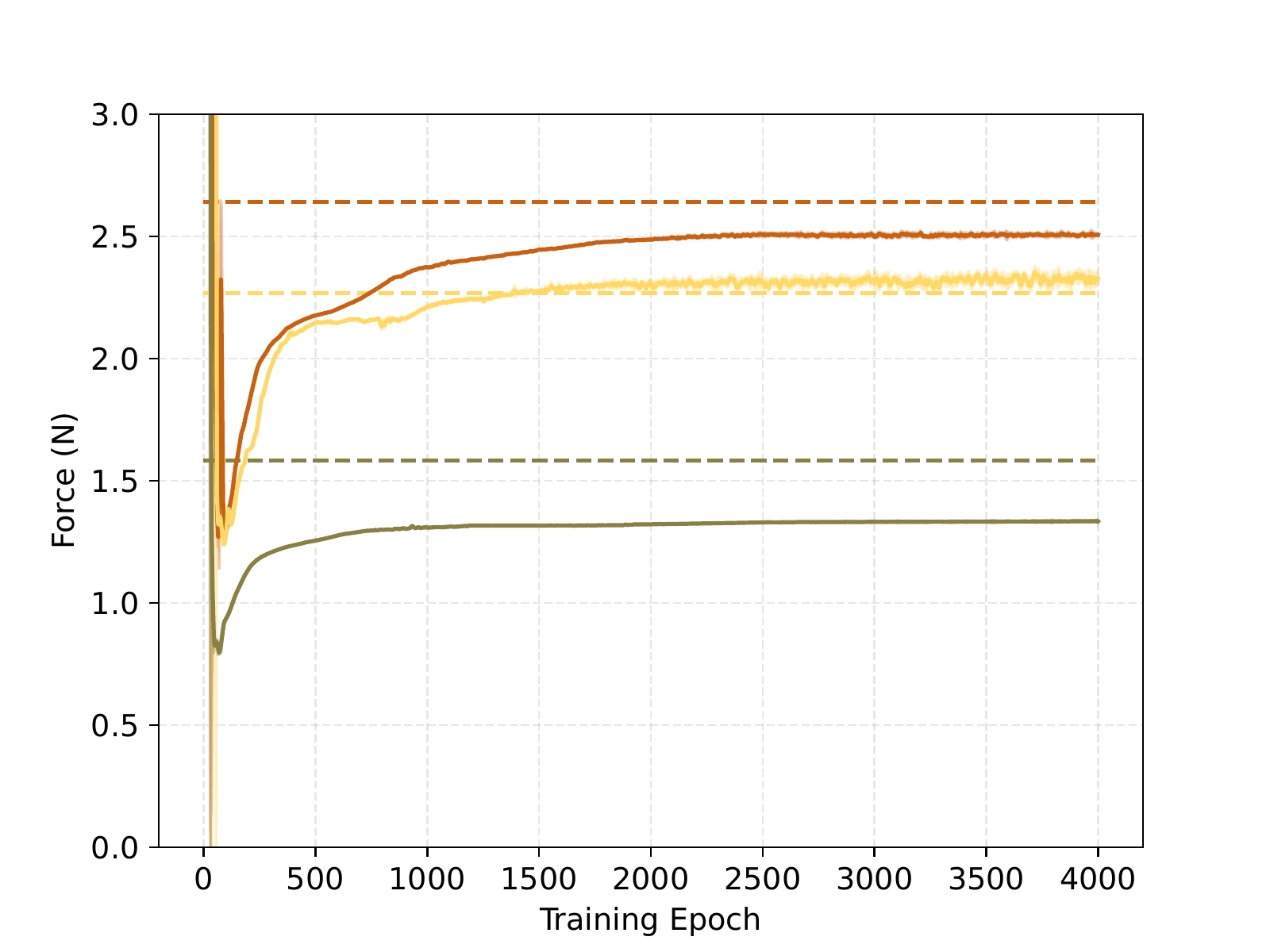}}
\vspace{-4mm}
\caption{Mean palm force during the training of imitation learning agents. Rolling window of $100$ epochs is used, $\pm1$ standard deviation is shaded. Dashed lines indicate mean palm force from demonstrations.}
\label{fig:palm_train}
\vspace{-4mm}
\end{figure*}

\section{Results and Discussion}
\label{sec:results_discussion}

The fingertip forces and palm forces applied by each demonstrator to the simulated environment with each combination of gripper and feedback condition is shown in Fig.~\ref{fig:force_demo}. As seen, fingertip force is significantly reduced ($p < 0.001$) in all conditions where fingertip force feedback is rendered to the demonstrator. Interestingly, without fingertip feedback some participants applied a greater fingertip force with the RUTH gripper than is possible with an average human hand~\cite{Bohannon2006ReferenceMeta-analysis}, whereas with fingertip force feedback all participants are well within expected ranges for human grasping. This indicates that the demonstrator is able to modulate their grasping force according to this feedback, only applying a grasp force that they perceive to be necessary for the task. Without fingertip force feedback, there are no sensory cues to indicate that a successful grasp has been made or that the demonstrator's grasp force is sufficient for a successful pick and place action. Palm force follows a similar trend, with statistical significance ($p < 0.001$) observed between conditions without palm force feedback (NFF and FFF) and conditions with palm force feedback (FPFF), with the exception of FFF and FPFF for the MANO hand. Results indicate that demonstrations under the FFF condition with the MANO hand also exhibit a reduced palm force compared to NFF ($p<0.05$). The main source of palm force is, qualitatively, due to normal reaction force from the desk in simulation. This could explain the reduction in palm force between NFF and FFF for the MANO hand; the fingertip force feedback may have transmitted some aspects of palm force. This is because the open-hand position of the MANO hand leaves fingertips approximately parallel to the surface of the desk, so fingertip force feedback can reflect palm force feedback. This is in contrast to the Franka Emika and RUTH grippers, where the gripper fingers are approximately perpendicular to the desk, and therefore will not transmit normal forces from desk contact.

Fig.~\ref{fig:fingertip_train} shows the mean fingertip force applied by imitation learning agents during training. As expected, all agents converge approximately to the mean fingertip force applied by their respective demonstrations, however NFF exhibits a much larger variance in fingertip force than other conditions. This is reflective of the large variance exhibited in each set of NFF demonstrations, with large interquartile ranges seen in Fig.~\ref{fig:fingertip_demo}. The palm force applied by imitation learning agents during training are shown in Fig.~\ref{fig:palm_train}. Surprisingly, these do not converge to their respective demonstration means, however, all conditions---with the exception of the RUTH hand under the FFF condition---fall approximately within the demonstration interquartile ranges seen in Fig.~\ref{fig:palm_demo}, indicating that the learnt trajectories are somewhat similar to their respective demonstrations. Importantly, conditions without palm force feedback converge to dramatically higher force values than conditions with palm force feedback, showing that exposing the demonstrator to force feedback allows an agent to learn a trajectory that exerts less force on its environment. In general, these fingertip and palm force results show that rendering force feedback to a demonstrator will produce safer, more stable agents after imitation learning, even though force is not inherently shown to the agent during training.

Beyond examining fingertip and palm force, we also inspect the average execution time of demonstration trajectories and learnt trajectories. Summarized in Table~\ref{table:execution_time}, it can be seen that trajectories demonstrated and trained under conditions with feedback have a faster execution time. This is expected; feedback allows the demonstrator to more accurately infer when a successful grasp has been made, allowing them to proceed with the pick-and-place task with minimal delays. Learnt trajectories are generally reflective of demonstration trajectories, in that faster demonstration trajectories correspond to faster learnt trajectories, however some discrepancies exist between demonstration and trained agent execution time. This is particularly true for FPFF results of the trained agent, which are considerably different to the average demonstration execution time. This highlights an interesting consequence of training an agent to complete a task on a wide range of demonstrations recorded from multiple experts; the average demonstration may not necessarily relate to the trajectory that is easiest to imitate. The average execution times of the Franka Emika gripper demonstration trajectories show far less improvement with feedback than the other grippers/hands. This may be because the 1 DoF gripper has only two fingers, meaning a more precise motion must be used to achieve a successful grasp, limiting how quickly demonstrators were able to perform the motion. Demonstration execution times for the RUTH hand, however, are much faster, due to the three-fingered nature of the gripper making grasping much more robust and easy to achieve with coarser motions. Finally, the MANO hand shows a dramatic improvement of both demonstration and trained agent execution time when fingertip and palm force feedback are utilized. This may be due to participants behaving more consistently under the FPFF condition, due to a higher level of immersion being achieved.

\begin{table}[t]
\caption{Summary of average trajectory execution times.}
\begin{tabular}{|c|c|c|c|c|c|c|}
\hline
\multirow{3}{*}{Gripper} & \multicolumn{6}{c|}{Average Execution Time (s)}                                                                                                   \\ \cline{2-7} 
& \multicolumn{3}{c|}{Demonstration} & \multicolumn{3}{c|}{Trained Agent} \\ \cline{2-7} 
                & {NFF} & {FFF} & {FPFF} & {NFF} & {FFF} & FPFF \\ \hline
Franka Emika    & {20.52} & {18.47} & \textbf{17.73} & {27.40} & {14.30} & \textbf{11.37}      \\ \hline
RUTH            & {15.53} & \textbf{11.94} & {12.69} & {18.19} & \textbf{11.75} & {16.60}      \\ \hline
MANO & {20.67} & {19.37} & \textbf{14.56} & {18.77} & {18.74} & \textbf{11.97}      \\ \hline
\end{tabular}
\label{table:execution_time}
\vspace{-6mm}
\end{table}

\section{Conclusions and Future Work}
\label{sec:conclusions}

In this work, we have explored the effect of two modes of force feedback on imitation learning. By utilizing a force feedback glove and a robot arm, we rendered fingertip-level and palm-level force feedback to human demonstrators, then used this immersive platform to collect demonstration trajectories of a pick-and-place task from multiple participants using three different grippers. Three feedback conditions were tested: no force feedback, fingertip force feedback, and fingertip and palm force feedback. Demonstration trajectories showed that demonstrations recorded with fingertip feedback applied significantly lower fingertip force to the simulated object, and demonstrations recorded with palm feedback exhibited significantly reduced palm force. These benefits were shown to translate to trained agents; the learnt trajectories resulted in far less force being applied to the grasped object and environment, indicating that force feedback may be the key to quality imitation learning where safety and force limits are critical. Furthermore, it was observed that feedback led to both demonstration and learnt trajectories that were quicker to execute, indicating that feedback may also be useful where real-time applications are desired. In summary, immersive demonstrations achieved via force feedback unlock safer, more execution-efficient imitation learning.

In future, a more complex range of tasks will be investigated with more complex environment state information, particularly those where feedback is vital for task completion (for example, when vision is significantly occluded). The relationship between demonstration quality and psychological embodiment will also be explored; virtual reality headset vs a fixed view camera, visual and auditory qualities of the simulated environment, and anthropomorphic vs non-anthropomorphic gripper design all contribute considerably to how immersive the experience is, which will in turn impact the ability of the expert to produce a quality demonstration. Finally, how well the trained agent can be transferred to a real robot will be studied to begin to attempt to realize dexterous manipulation via imitation learning in real world settings.

\section*{Acknowledgement}

The authors would like to thank the Robot Intelligence Lab, the Morph Lab, the SiMMS Research Group, and Jinxin Liu for providing equipment used in this work.

\addtolength{\textheight}{-5.0cm}  


\bibliographystyle{IEEEtran}
\bibliography{references}

\end{document}